\documentclass{llncs}
\usepackage{amsmath}
\usepackage{mathtools}
\usepackage{amssymb}
\DeclareMathOperator*{\argmin}{argmin}
\usepackage{times}
\usepackage{epsfig}
\usepackage{graphicx}
\usepackage{amsmath}
\usepackage{amssymb}
\usepackage{subfig}

\usepackage{amsthm}
\usepackage{algorithm}
\usepackage{algorithmicx}
\usepackage{algpseudocode}
\usepackage[nolist,nohyperlinks]{acronym}
\usepackage{color}
\usepackage{bbm}
\usepackage{bold-extra}

\usepackage{makeidx}  

\begin{document}

\title{Towards Learning free Naive Bayes Nearest Neighbor-based Domain Adaptation}
\titlerunning{ICIAP Paper}  
%
\author{Faraz Saeedan\inst{1} \and Barbara Caputo\inst{1,2}}
\authorrunning{Faraz Saedaan et al.} 
%
\tocauthor{Faraz Saeedan and Barbara Caputo}
\institute{Sapienza University, Rome, Italy,\\
\email{caputo@dis.uniroma1.it},
\and
Idiap Research Institute, Martigny, Switzerland}

\maketitle              

\begin{abstract}
As of today, object categorization algorithms are not able to achieve the level of robustness and generality necessary to work reliably in the real world. Even the most powerful convolutional neural network we can train fails to perform satisfactorily when trained and tested on data from different databases.
This issue, known as domain adaptation and/or dataset bias in the literature, is due to
a distribution mismatch between data collections. Methods addressing it go from 
max-margin classifiers to learning how to modify the features and obtain a 
more robust representation. Recent work showed that by casting the problem into the image-to-class recognition framework, the domain adaptation problem is significantly alleviated \cite{danbnn}. Here we follow this approach, and show how a very simple, learning free Naive Bayes Nearest Neighbor (NBNN)-based domain adaptation algorithm can significantly alleviate the distribution mismatch among source and target data, especially when the number of classes and the number of sources grow. Experiments on standard benchmarks used in the literature show that our approach (a) is competitive with the current state of the art on small scale problems, and (b) achieves the current state of the art as the number of classes and sources grows, with minimal computational requirements.

\end{abstract}
\section{Introduction}

In the last years the computer vision research community's
attention has been driven towards the existence of differences across predefined 
image datasets and the necessity to recompose these idiosyncrasies. The main reason behind this need is the increasing amount of available image data 
sources and the absence of a unique general learning method that can 
perform well across all of them. In practice training a classifier on a dataset
(e.g. Flicker photos) and testing on another (e.g. images captured
with a mobile phone) produces very poor results although the task (i.e. the set
of depicted object categories) is the same.

In this context the notion of \emph{domain} already used in machine learning 
for speech and language processing has been extended to visual problems. 
A source domain ($S$) usually contains a large amount of labeled images, while a
target domain ($T$) refers broadly to a dataset that is assumed to have different
characteristics from the source, and few or no labeled samples. 
Formally we can say that two domains differ when for their probability distributions 
it holds $P_S(x,y)\neq P_T(x,y)$, where $x\in\mathcal{X}$ 
indicates the generic image sample and $y\in\mathcal{Y}$ the corresponding class label.
Specific annotator tendencies may influence the conditional distributions 
implying $P_S(y|x)\neq P_T(y|x)$. Other typical causes of visual domain 
shift include changing in the acquisition device, image resolution, lighting, background, 
viewpoint and post-processing \cite{bias}. Most of these information are directly encoded in 
the descriptor space $\mathcal{X}$ chosen to represent the images and may induce a difference among 
the marginal distributions $P_S(x) \neq P_T(x)$.

In 2013, Tommasi and Caputo showed that by casting the domain adaptation problem into the Naive Bayes Nearest Neighbor framework one could achieve a very high level of generalization, thanks to the intrinsic properties of NBNN classifiers \cite{danbnn}.
The proposed approach used distance metric learning to leverage over the source knowledge at the local patch level. This brought strong results in the semi-supervised and unsupervised domain adaptation scenarios, but the method is computationally expensive and thus not suitable to work on real-time systems, like smatphones or robots. 

Here we propose a simple, learning free domain adaptation method that makes it possible to exploit the generalization power of NBNN in the domain adaptation setting. We leverage over the source patches by randomly selecting a subset of them, and adding them to the target patches. To further increase the descriptive power of the descriptors, we perform data augmentation both on the source and the target data, as it is standard practice in the Convolutional Neural Network literature \cite{augmentation}. The combined effect of these two simple actions is remarkable: on commonly used benchmark databases, our approach is on par with the current state of the art when there is a single source from which to adapt, and when the number of classes is limited. In the more challenging and more realistic settings of multiple sources and increasing number of classes, our algorithm achieves the state of the art.  

The rest of the paper is organized as follows: after reviewing previous work (section \ref{rel-work}) we revise the basic definitions for domain adaptation (section \ref{da}) and the NBNN framework (section \ref{nbnn}). Section \ref{rand-da-nbnn} introduces our approach, while section \ref{experiments} presents its thorough experimental evaluation. We conclude with a summary discussion and outlining possible future avenues for research.


%
\section{Related Work}
\label{rel-work}
The problem of domain adaptation stems from the fact that supervised learning methods fail to generalize across datasets \cite{bias}. Although this problem exists in various applications \cite{bendavid,quionero,daume,mcdonald}, the visual recognition community has just recently shown interest in dealing with it \cite{gopalan,bergamo,fritz}. Failure to generalize across datasets has been attributed to the mismatch among various characteristics of the considered databases, and is usually referred to as the `dataset bias' problem \cite{bias}. The fact that different image datasets vary considerably in quality, point of view and image contents, reveals that addressing the domain adaptation problem can significantly improve the performance of visual recognition applications.

Several approaches have been adopted for reducing the distance between datasets.
These approaches vary from transferring source data to the target domain \cite{fritz} or transferring both source and target to a third space\cite{gopalan}. Unfortunately, despite all efforts, \cite{landmark} showed that currently existing selective transfers do not offer significant improvement over random transfers. As an alternative to the enrichment of the target data through instance based transfer from the source, attempts have been made to modify the classifier in order to resolve the mismatch problem \cite{duan,bruzzone,yang,khosla} during training.

While the image to image paradigm is the dominant approach in the above-mentioned methods, \cite{Q1,Q2} suggested that one can replace NBNN for Bag Of Words (BOW) combined with an image-to-image classification paradigm, in favor of an image-to-class recognition framework. This idea helps release the domain transfer from the known shortcomings of BOW representations \cite{boiman}.

Even though NBNN has been tested on several visual learning applications, the use of this classification paradigm in domain adaptation has been limited. Only in 2013 \cite{danbnn} exploited its potential in a metric learning approach, and showed that using NBNN, one can easily surpass the state of the art among BOW-based algorithms presented so far. Be that as it may, the possible usages of this method, called DA-NBNN,  are restricted due to the computational complexity. Indeed, once the amount of classes, the number of sources and the number of data for each class and source grow, using DA-NBNN becomes computationally prohibitive. 
Our approach overcomes these computational limitations while preserving, and often significantly surpassing, the performances of DA-NBNN, proposing the first learning free NBNN-based domain adaptation method in the literature.

%
\section{Problem Setting and Definitions}
\label{prob-setting}
In this section we set the scene by introducing formal definitions for the domain adaptation problem (section \ref{da}) and the NBNN classification framework \ref{nbnn}. The notion introduced in this section will then be used to present our algorithm.

\subsection{Domain Adaptation}
\label{da}
Domain Adaptation is the problem where knowledge from the source domain $\mathcal{D}^s$ is used to enrich and hence improve the performance in the target domain $\mathcal{D}^t$. This knowledge from the souce might be in the form of instances or data, or model parameters, or metric induced by the source. It is usually implicitly assumed that the labeled data on the target domain does not exist (unsupervised setting) or it is scarce (semi-supervised setting). Although the source and the target domains are different, they use equal label sets \cite{musket} $\mathcal{Y}^s=\mathcal{Y}^t$.\\
The core cause of mismatch between the two domains is attributed to the difference in the distribution of these labels. The conditional probability of labels given features are not completely coincident $P^s(Y|X) \sim P^t(Y|X)$ and the marginal data distributions are not equal either $P^s(X)\neq P^t(X)$. In this paper, we will focus exclusively on the semi-supervised setting.  

\subsection{Naive Bayes Nearest Neighbor}
\label{nbnn}
In the Naive Bayes Nearest Neighbor (NBNN) classification framework, it is assumed that for each class there exists a distribution from which local descriptors are drawn independently of one another. This leads to the use of a Naive Bayes maximum a posteriori classifier \cite{boiman} where each feature $m$ votes for one of the classes in $c=\{1,...,C\}$. This voting is realized using the local distance between each feature and its nearest neighbor in class c. $Df2C(m,c)=||f_m-f_m^c||$. The generalization of this distance concept to image to class distance is straightforward:$D_I2C(F_i,c)=\sum{m=1}{M_i}D_f2C(m,c)$.\\
The output of the classifier would then be 
\begin{equation}
p=\argmin_c D_I2C(F_i,c)
\end{equation}
The distance to this optimum class p is called the positive distance while the distances to the rest of the classes $n:\{c\neq p\},
D_I2c(F_i,n)$ are called the negative distances.

\section{Learning free NBNN-based domain adaptation}
\begin{figure}[tb]
\centering
\includegraphics[width=99mm]{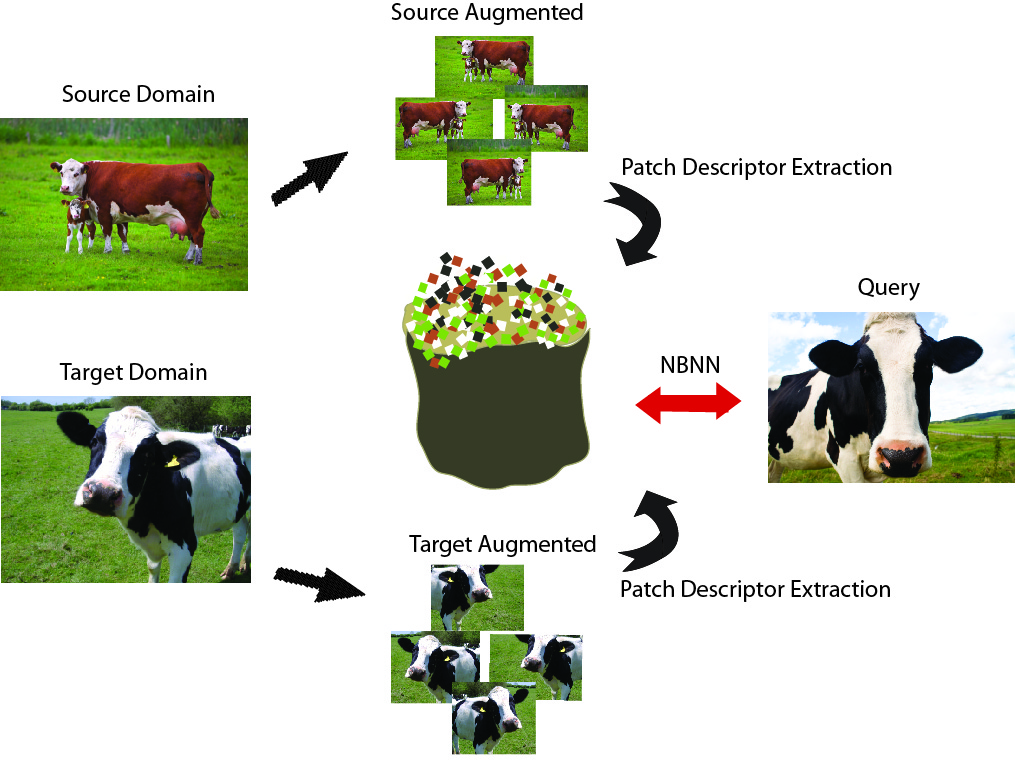}
\label{diagram}
\caption{An overview of our proposed learning free, NBNN-based domain adaptation approach for the class`cow': after performing data augmentation on both the source and target data, patches-based features are extracted from both, and a new target data set is created by merging the whole patches-based features extracted from the target with a fraction of those of the source, randomly selected from the whole sample data. This new pool of patches-based features is then used to build an NBNN classifier in the target domain.}
\end{figure}
\label{rand-da-nbnn}
As outlined above,
the problem of domain adaptation emerges when the training data for
the target task is scarce. 
Should it not be the case, any supervised learning algorithm would be  capable of learning a classifier, according to its learning abilities.
 It is also assumed that there exists at least another dataset with enough samples 
 to learn a good classifier (the source), 
 but since the two datasets have been acquired in two different domains, the performance obtained training on the source and testing on the target is not satisfactory.
 
The NBNN algorithm builds support sets for each class made of the collection of all the features extracted from patches of each of the training examples. Due to the scarcity of the data on the target, the support sets that can be built solely using features from the target samples will not contain enough features to guarantee a solid performance. In order to enrich these support sets,  \emph{our proposal is to use features extracted from the patches of the source images}.

How to select such patches-based features? In \cite{landmark}, the authors investigate a domain adaptation approach based on the idea of landmark samples from the source domain, which are relevant for the modeling of the target classifier. Although their approach is theoretically sound, experiments show that the learning method proposed to select such landmark is often statistically on par, and otherwise within a two percent range of performance, with a random selection of the learning samples. Motivated by this result, we apply the same philosophy  here to the patches-based features, and we propose to achieve domain adaptation in an NBNN-based framework by randomly sampling a percentage of the patches-based features from the source, adding them to the patches-based features of the target. We will show with experiments in the next section that this extremely simple and learning free strategy achieves amazingly good results on standard domain adaptation benchmark databases, while being reasonably stable with respect to the amount of features to be samples.

To further improve performance in our approach, we have tested the effect of performing data augmentation on the source and target data.
Data augmentation is a technique that, since the spectacular success of convolutional neural network in the visual classification arena, has been shown to be very effective in general for any classification algorithm \cite{augmentation}. Again, our experiments confirm the effectiveness of this strategy, even more so combined with the instance-based domain adaptation approach based on random sampling of patches-based features from the source. 
A schematic representation of the overall approach for the class `cow' is given in figure \ref{rand-da-nbnn}.
Note that adding the data augmentation step to our overall approach does not significantly increase the almost non-existent computational load in training.
This characteristic, combined with the remarkably good performances achieved especially as the number of classes and sources grow, makes our approach potentially attractive for applications where computational complexity should be low, like mobile robot or online, wearable systems. To the best of our knowledge, there are no previous instance-based, NBNN-based domain adaptation methods in the literature, nor the random sampling strategy has been ever tested in the NBNN learning framework for any learning to learn approach.

\section{Experiments}
\label{experiments}
In this section we describe the experiments we performed to assess our approach. We first describe the data, features and experimental setup used (section \ref{Datasets}), then we report the results obtained (section \ref{results}). We discuss our findings in section \ref{discussion}.

\subsection{Datasets, Features and setup}
\label{Datasets}

\subsubsection{Datasets}

We used the Office dataset, 
the standard test bed in domain adaptation which addresses the problem of object categorization between any two datasets of objects usually found in offices \cite{fritz}. This test bed consists of three domains namely Amazon, Webcam and Dslr. The Amazon dataset contains images obtained from online merchants. The images are centered and usually on a white background. Webcam and Dslr are respectively low resolution and high resolution images obtained from web cam and SLR cameras.  Unlike Amazon, they could be subject to various environmental disturbances such as lighting or background changes. The Office dataset contain 31 classes of images for each domain.\\
Having chosen 10 of the original 31 classes from office, \cite{geodesic} suggested that we can add images of the same 10 classes from Caltech-256 \cite{caltech} and form the Office+Caltech test bed in order to add a fourth domain in the office dataset.\\

\subsubsection{Features}

Following the protocol of \cite{danbnn}, images were all resized to a common width (256px) and then converted to grayscale. SURF features were extracted according to \cite{surf}. The final result was a set of features of length 64 that were consequently fed to a 1-nearest neighbor classifier. \\
The effect of data augmentation on both domains has also been studied. 
To this end, we have duplicated the exact procedure suggested in \cite{augmentation} and each image is converted into 10 images through the procedure of cropping and flipping.\\

\subsubsection{Setup}

Different pairs of datasets are chosen to act as the source and the target from the Office+Caltech group. From the source dataset, 20 images were selected to represent the source data but only 3 were chosen from the target in every class. When the target was Webcam, 15 images were selected instead of 20 as described in \cite{danbnn}. At this stage, since the Dslr dataset behaves very similarly to Webcam and it contains a lower number of images, we decided not to include it in our benchmarking. \\
The same sample selection protocol has been adopted for the 31 class adaptation experiments. 
The third setup that we considered is domain adaptation from more than one source with one target. To this end, all possible combinations of two sources to one target have been examined and benchmarked against the existing reported results in the literature.

\subsection{Results}
\label{results}
The first set of experiments was done on a subset of Office+Caltech consisting of 10 classes as explained in \cite{danbnn}.  
Figure \ref{fig:accuracies} shows the results in comparison to the state of the art and some baseline algorithms.\\

\newlength{\subfiglength}

\begin{figure*}[!t]
  \caption{Results for the 10 class experiments. Figure \ref{fig:acc_caltech256} shows the overall results obtained by our method compared against state of the art algorithms. Figure \ref{fig:acc_caltech256_classemes} shows the change in recognition rate on the Amazon-Caltech experiment of our method as the percentage of source data transferred to the target set increases, for the cases no augmentation, only source data augmentation, only target data augmentation and both source and target data augmentation. Analogous results are shown in figure \ref{fig:acc_caltech256_object_bank} and figure \ref{fig:acc_sun09} for the Webcam-Amazon and Caltech-Webcam settings respectively.}

  \centering
  \setlength{\subfiglength}{4.5cm}
  \subfloat[10-class experiments, Overall]{%
    \includegraphics[height=\subfiglength]{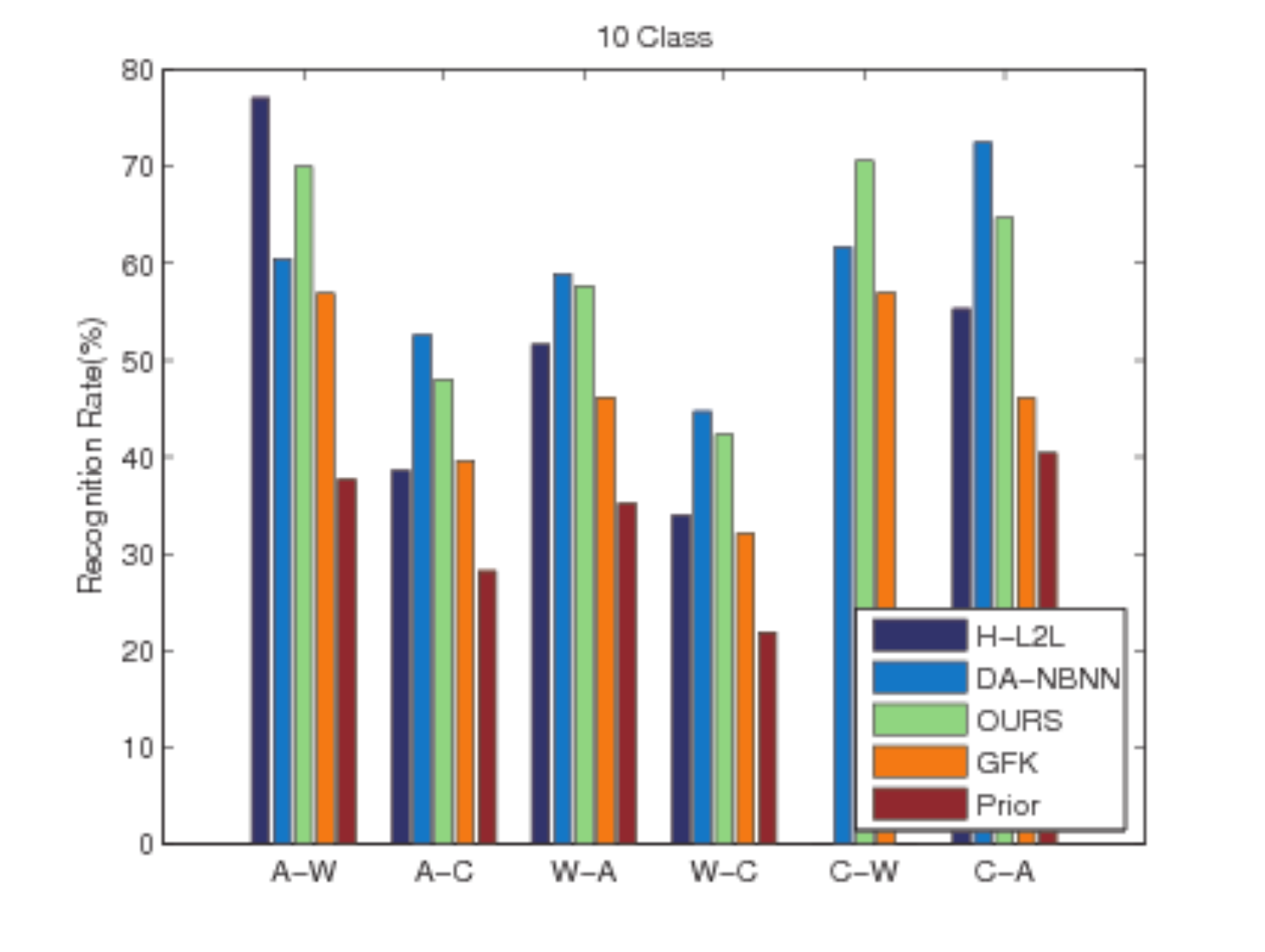}
    \label{fig:acc_caltech256}
  }%
  \subfloat[A-C Results]{%
    \includegraphics[height=\subfiglength]{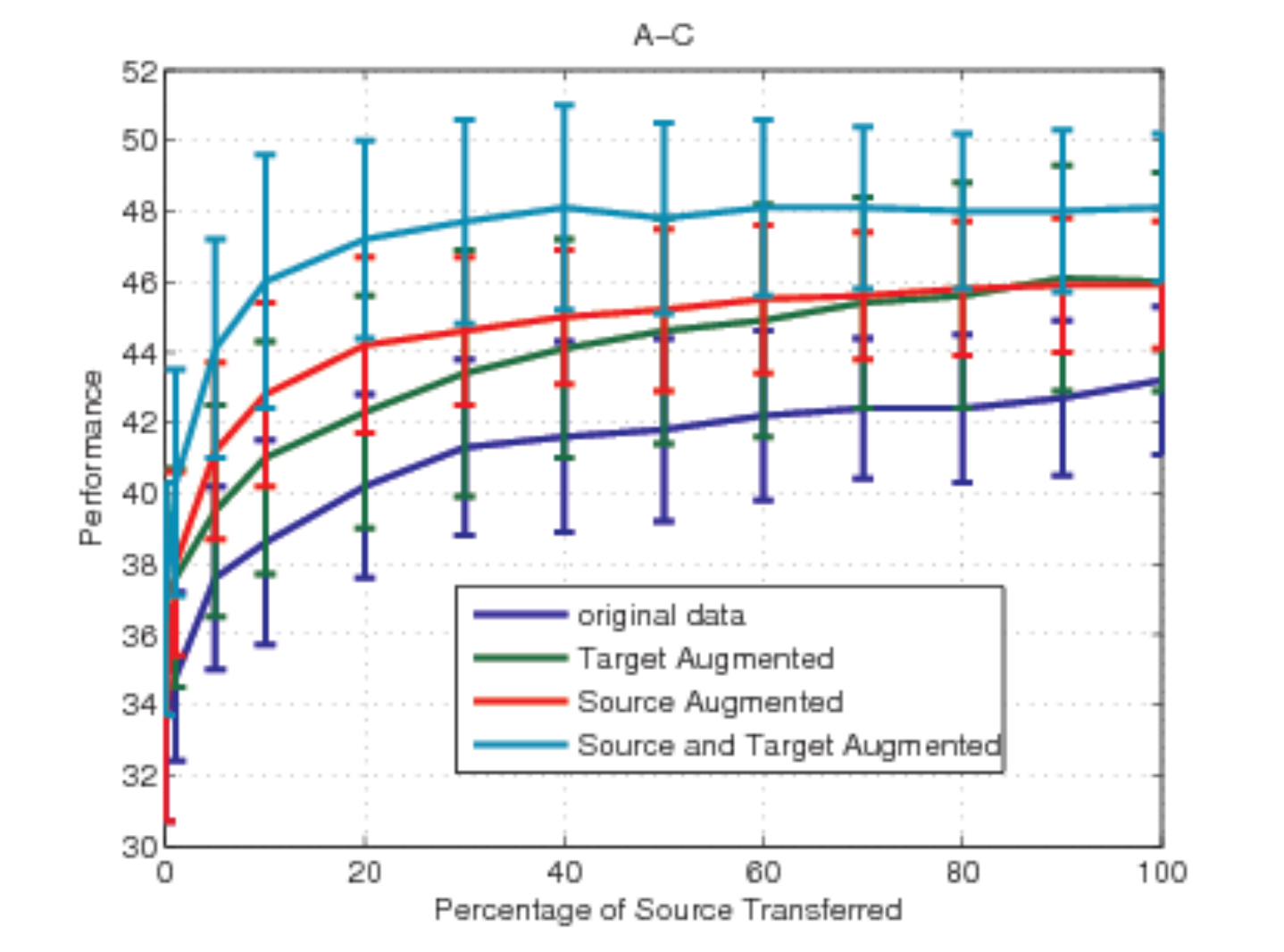}
    \label{fig:acc_caltech256_classemes}
  }\\
  \vspace{-0.2cm}
  \subfloat[W-A Results] {%
    \includegraphics[height=\subfiglength]{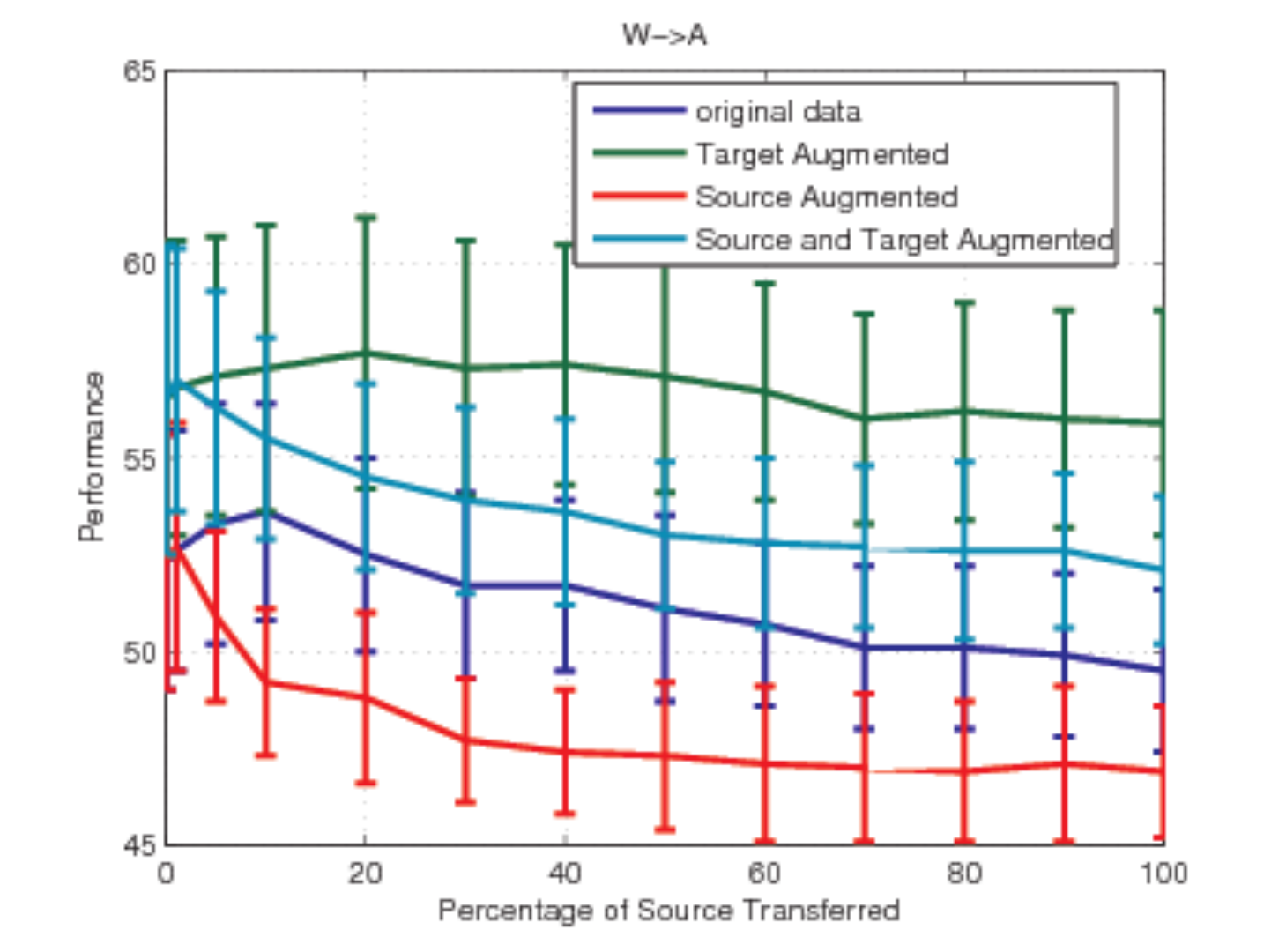}
    \label{fig:acc_caltech256_object_bank}
  }%
  \subfloat[C-W Results] {%
    \includegraphics[height=\subfiglength]{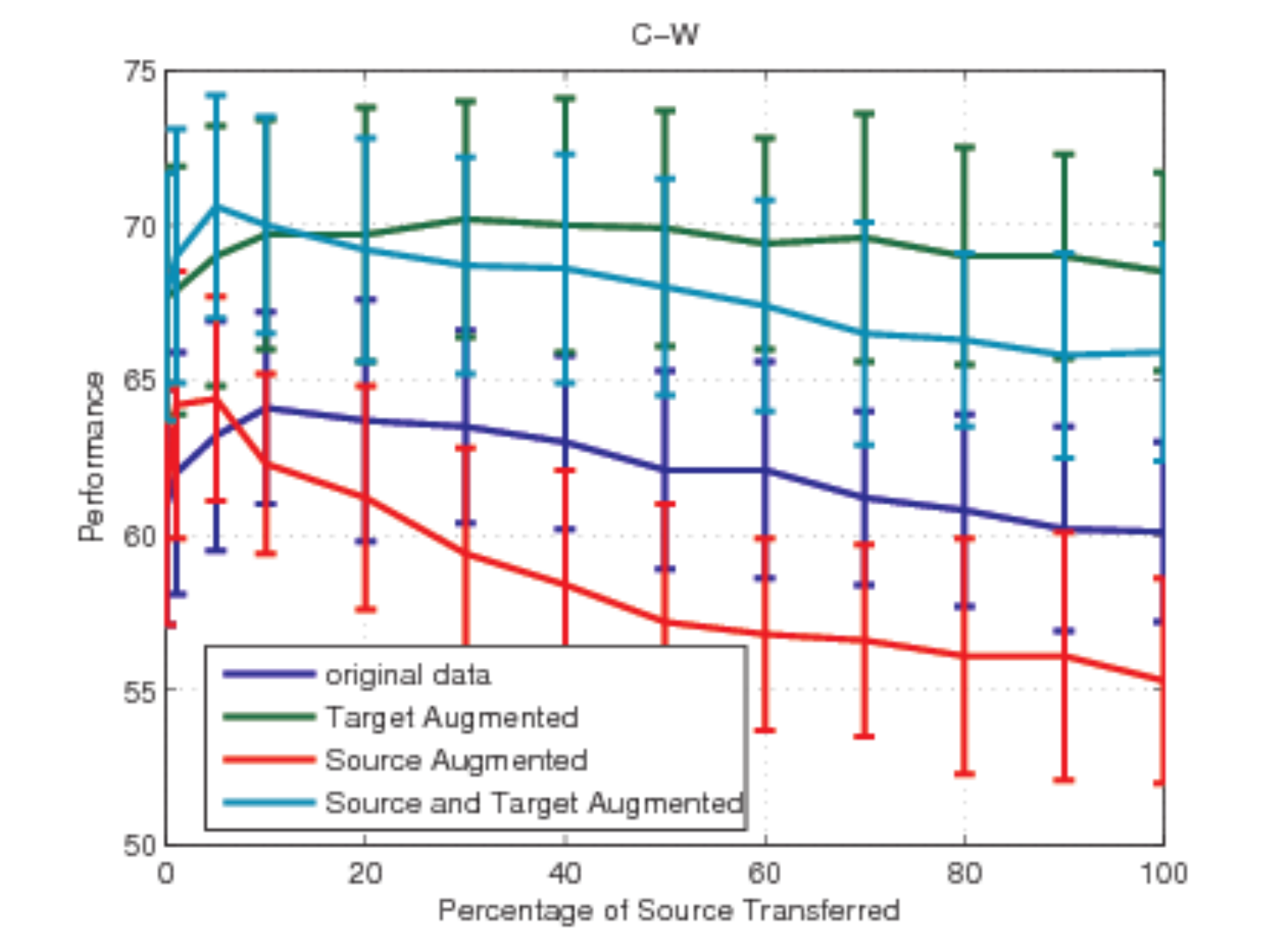}
    \label{fig:acc_sun09}
  }\\
\label{fig:accuracies}
\end{figure*}





Figures~\ref{fig:acc_caltech256_classemes}, \ref{fig:acc_caltech256_object_bank} and \ref{fig:acc_sun09}  show the changes in the recognition rate with the increase of the percentage of descriptors, randomly transferred to the target from the source. For a better understanding of the effects of different factors, four cases have been demonstrated together. Original data is where there is no augmentation done neither on the target nor on the source domains. The cases where only the source and only the target domains have been augmented are referred to as Source augmented and Target Augmented respectively. Source and Target Augmented is where both domains have been over-sampled.\\







The second set of experiments is done on the 31 class Office dataset. The experiments are done exactly inline with what explained and done in \cite{danbnn}. Table \ref{31 class} shows the results with comparison to the state of the art both using NBNN and the state of the art based on a method other than NBNN. Some further baselines are also included for better comparison.\\

\begin{table}[htb]

\centering
\begin{tabular}{|c|| c| c| c|}
\hline

Algorithm & $A\longrightarrow W$ & $W\longrightarrow D$ & $D\longrightarrow W$\\
\hline
\hline
BOW&$34.9\pm 0.6$&$48.9\pm 0.5$&$38.4\pm 0.4$\\
\hline
GFK&$46.4\pm 0.5$&$66.3\pm 0.4$&$61.3\pm 0.4$\\
\hline
NBNN&$40.0\pm 2.0$&$67.2\pm 2.5$&$70.7\pm 1.2$\\
\hline
I2CDML&$47.9\pm 1.3$&$72.8\pm 2.1$&$73.8\pm 1.6$\\
\hline
$H-L2L(hp-\beta)$&$\mathbf{76.2\pm 0.02}$&$67.8\pm 0.05$&$66.0\pm0.01$\\
\hline
DA-NBNN&$52.8\pm 3.7$ & $76.2\pm 2.5$&$76.6\pm1.7$\\
\hline
OURS&$55.0\pm3.3$&$\mathbf{77.5\pm2.0}$&$\mathbf{78.2\pm1.4}$\\
\hline
\end{tabular}
\caption{31 class Office dataset experiments, semi-supervised setting}
\label{31 class}
\end{table}

\begin{figure}[h!]
\centering
\includegraphics[width=99mm]{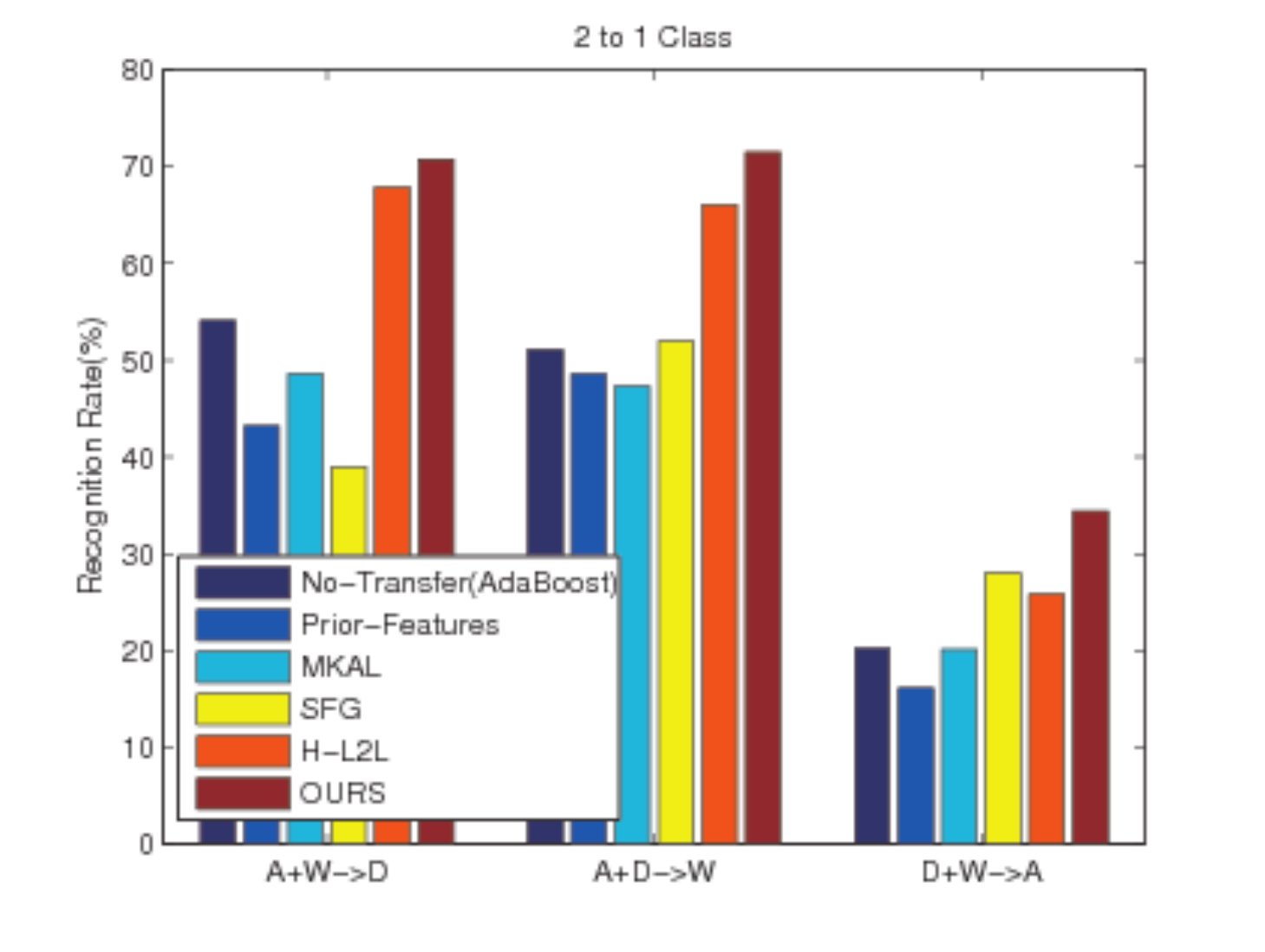}

\caption{Accuracy on target domains with multiple sources (A:Amazon, W:Webcam, D:DSLR), 31 class, semi-supervised}
\label{2to1}
\end{figure}

The Third and last set of experiments are those run using more than one source domain. The Results can be seen in Figure \ref{2to1}. Not all Algorithms can be extended to cover the case of several sources and so only those who had this advantage were included in the comparison. For the experiments the exact test set of \cite{l2l} has been used.

\subsection{Discussion}
\label{discussion}
The biggest advantage of our proposed method is its simplicity
combined with its strong performance over growing number of classes and source domains.
 It also performs surprisingly well in comparison to other algorithms. The results in Figure \ref{fig:accuracies} show that while different algorithms have varying performances on various test settings,
 our method is never worse than the second best. In particular, compared to DA-NBNN \cite{danbnn} (which is the state of the art among all the methods that exploit an NBNN approach), our method outperforms it in 2 cases (A-W and C-W), while DA-NBNN performs better in two cases (C-A and A-C). In the remaining two cases (W-A, W-C) their performance is close. In fact, the $p$ test shows that in these two experiments there is no statistical evidence of superiority for either of the algorithms. 

Our method performs significantly better than L2L \cite{l2l} where L2L is the state of the art among methods that do not use NBNN. In four of the experiments, L2L achieves inferior results than ours, while only in one setting shows superiority. Note that the accuracy values reported for L2L  have been taken from \cite{l2l}, where no result was reported for the C-W experiment.

Using the 31 class Office setting, one can study and compare the scalability of the algorithms with respect to the number of classes. Addressing this type of scalability for our method appears very straightforward. The fact that there is no training, makes things very easy and faster. Table \ref{31 class} shows that, performance-wise, our method scores higher than DA-NBNN in all three experiments and better than L2L in two out of three cases. 

Figure \ref{2to1} Compares the recognition rate for all possible combinations of two sources and one target in the Office dataset. For DA-NBNN it is not clear how it could be extended to this case and no experiments of the kind have been reported by its authors. L2L supports this case and it has been included in the benchmark. It can be seen that our method outperforms all the others for all three cases of experiments.

An open issue in our method is of course which percentage of the source data should be randomly selected and then added to the target data, in relation to the data augmentation procedure. Results shown in figures \ref{fig:acc_caltech256_classemes}-\ref{fig:acc_sun09} show that 
in general the combination of source plus target data augmentation and random sampling of  around 20\% of patches-based features from the source seems to achieve strong performance, always better than the original data. Still, as it can be seen from the figures, the actual optimal performance might vary in terms of percentage of sampling and/or data augmentation strategy for different settings. Although accuracy results are on average quite stable, and therefore the algorithm could be used in online systems even in its current form with good expectations about performance, it would be desirable to explore further the issue of the data selection and find principled ways of selecting the patches to transfer from the source to the target so to have guarantees about the optimality of the procedure. Of course, that would come at the expenses of the current negligible computational cost of the approach.

\section{Conclusions}

The contribution of this paper is a learning free Naive Bayes Nearest Neighbor based domain adaptation method that is competitive with the current state of the art on the standard Office-Clatech benchmark database, and that achieves the state of the art when the number of classes and sources grows. The method consists in performing a random selection of patches-based local features from the source to the target, combined with a data augmentation strategy mutated from the CNN literature. The resulting algorithm is extremely simple but also remarkably effective, especially when the number of classes and sources grows. An open challenge is how to select the best percentage of source data to add to the target: even though our experimental evaluation indicates that as a rule of thumb sampling around twenty percent of the overall sample data (i.e. after data augmentation) in general leads to very good results, future work will focus on how to determine how much to sample in a principled manner, while at the same time not increasing excessively the computational cost of the approach.

%
\newpage

\end{document}